\documentclass[10pt,twocolumn,letterpaper]{article}

\usepackage[pagenumbers]{wacv} 

\usepackage{graphicx}
\usepackage{amsmath}
\usepackage{amssymb}
\usepackage{booktabs}

\usepackage{ulem}
\usepackage{multirow}
\usepackage{url}
\usepackage{makecell}

\newcolumntype{M}[1]{>{\centering\arraybackslash}m{#1}}

\newcolumntype{?}{!{\vrule width 1pt}}

\usepackage[pagebackref,breaklinks,colorlinks]{hyperref}

\usepackage[capitalize]{cleveref}
\crefname{section}{Sec.}{Secs.}
\Crefname{section}{Section}{Sections}
\Crefname{table}{Table}{Tables}
\crefname{table}{Tab.}{Tabs.}

\def\wacvPaperID{1658} 
\def\confName{WACV}
\def\confYear{2025}

\begin{document}

\title{FlashMix: Fast Map-Free LiDAR Localization via Feature Mixing \\ and Contrastive-Constrained Accelerated Training}

\author{Raktim Gautam Goswami$^{1}$, Naman Patel$^{1}$, Prashanth Krishnamurthy$^{1}$, Farshad Khorrami$^{1}$
\thanks{$^{1}$Control/Robotics Research Laboratory (CRRL), Department of Electrical and Computer Engineering, NYU Tandon School of Engineering, Brooklyn, NY, 11201. E-mails: {\tt\small \{rgg9769,nkp269, prashanth.krishnamurthy,khorrami\}@nyu.edu}. 
This work was supported in part by ARO under Grant W911NF-22-1-0028 and in part by the New York University Abu Dhabi (NYUAD) Center for Artificial Intelligence and Robotics (CAIR), funded by Tamkeen through NYUAD Research Institute Award under Grant CG010.
}
}
\maketitle
\begin{abstract}
Map-free LiDAR localization systems accurately localize within known environments by predicting sensor position and orientation directly from raw point clouds, eliminating the need for large maps and descriptors. However, their long training times hinder rapid adaptation to new environments. To address this, we propose FlashMix, which uses a frozen, scene-agnostic backbone to extract local point descriptors, aggregated with an MLP mixer to predict sensor pose. A buffer of local descriptors is used to accelerate training by orders of magnitude, combined with metric learning or contrastive loss regularization of aggregated descriptors to improve performance and convergence. We evaluate FlashMix on various LiDAR localization benchmarks, examining different regularizations and aggregators, demonstrating its effectiveness for rapid and accurate LiDAR localization in real-world scenarios. The code is available at \href{https://github.com/raktimgg/FlashMix}{https://github.com/raktimgg/FlashMix}.
\end{abstract}

\section{Introduction}
Localization systems form the backbone of many modern technologies, from navigation to autonomous driving. These systems rely on sensors like LiDARs and cameras to determine an agent's position and orientation in a scene. LiDARs often prove more reliable, particularly in environments where appearances fluctuate. Conventional approaches for LiDAR localization use place recognition algorithms to retrieve a target point cloud from a database and perform registration to ascertain the query's pose~\cite{scancontext, pointnetvlad, salsa, spectralgv}. However, this strategy necessitates significant memory for storing map points and descriptors in addition to computationally intensive registration processes. 

Recently, map-free LiDAR localization systems have shown great promise for pose estimation in known environments. Initially developed for camera images, these systems aim to predict sensor pose directly through regression, potentially reducing the need for memory-intensive maps and descriptors. These approaches either directly predict 6-DoF pose or estimate scene coordinates to determine pose using a Perspective-n-Point solver \cite{pnp} within a Random Sample Consensus (RANSAC) \cite{ransac} framework. While effective for small, camera-captured scenes, these methods face challenges when scaling to large-scale environments.

LiDAR-based map-free localization approaches were subsequently introduced to capitalize on the rich geometric information provided by LiDAR sensors~\cite{pointloc}. Further developments~\cite{nidaloc, posepn++, sgloc, hypliloc} demonstrated the ability to achieve low localization errors in diverse environments through improvements in training methodologies and architectures. Despite promising results across various LiDAR datasets, these approaches face a significant challenge: current pose regression networks typically require lengthy training periods, often lasting hours to days, due to the need for individual training in each scene. This limitation hinders their practicality for robotics applications such as navigation and manipulation, which rely on rapid adaptation to new scenes for subsequent tasks.
\begin{figure}
    \centering
    \includegraphics[width=\linewidth]{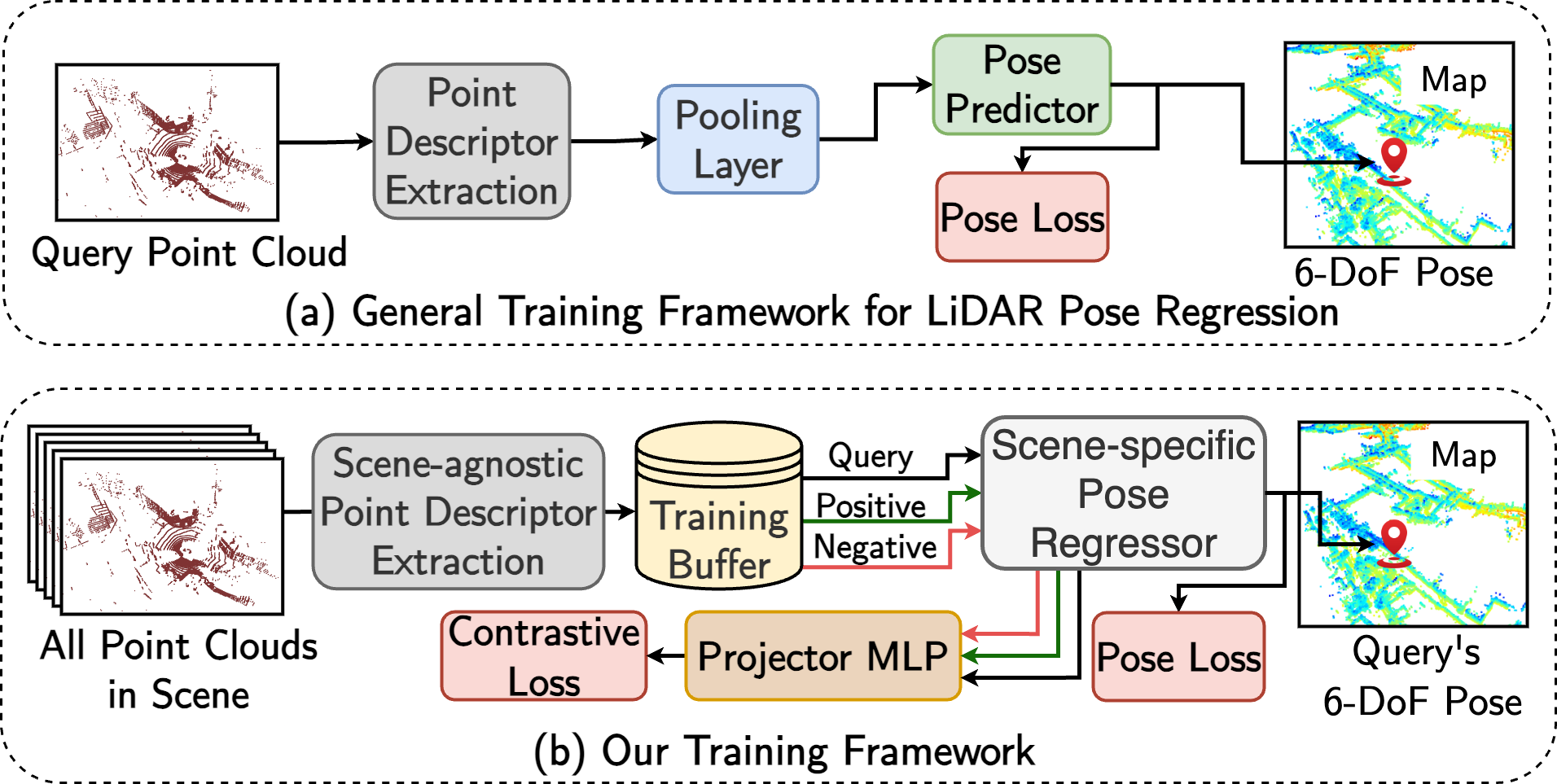}
    \caption{Comparision of LiDAR pose regression-based framework (top) with our fast map-free LiDAR localization system.}
    \label{fig:flow1}
    \vspace{-0.2in}
\end{figure}
\begin{figure*}
    \centering
    \includegraphics[width=0.9\linewidth]{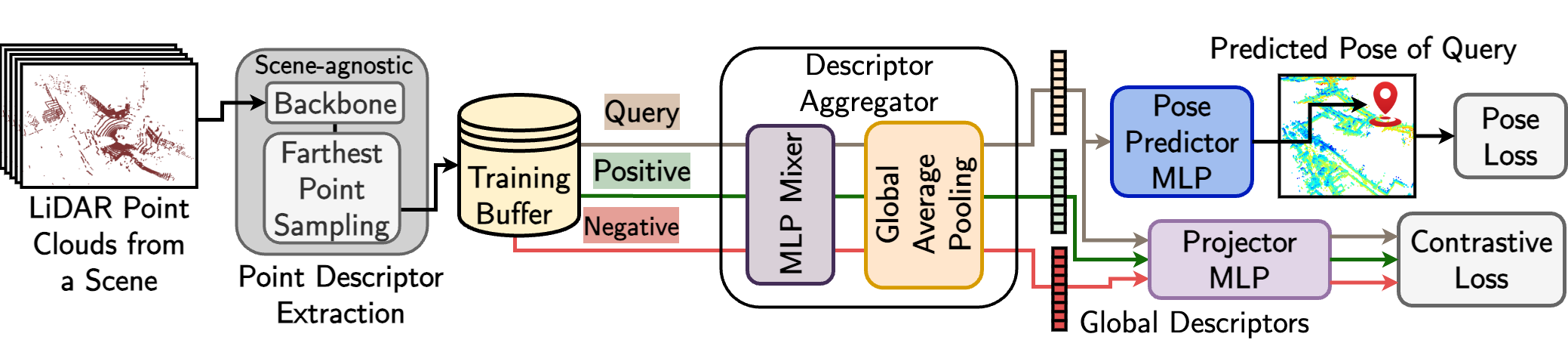}
    \caption{FlashMix framework: A scene-agnostic backbone extracts local descriptors from farthest point sampled point clouds to store in a training buffer. An MLP Mixer and global average pooled aggregate descriptor predicts pose from trained pose and contrastive loss. }
    \label{fig:flow2}
\end{figure*}

PosePN++~\cite{posepn++} introduced the concept of a universal feature encoder, showing that encoder weights trained on one scene could be transferred to another. While this approach reduced training time by requiring only decoder retraining for new scenes, it still faced limitations in quickly adapting to diverse environments without compromising accuracy. FlashMix, shown in Fig.~\ref{fig:flow1}, directly addresses these challenges, offering a solution that enables rapid adaptation to new scenes while maintaining high localization accuracy. By combining a pre-trained backbone with a scene-specific regressor and innovative training techniques, FlashMix significantly reduces training times for LiDAR localization systems without sacrificing performance. A pre-trained, scene-agnostic backbone is used to extract local point descriptors, which are then aggregated using an MLP mixer to estimate sensor pose. FlashMix creates a training buffer by extracting local descriptors for each point in the point cloud. This buffer is used to train a descriptor aggregator and a pose predictor, which together form the scene-specific pose regressor. This approach, shown in Fig.~\ref{fig:flow2}, accelerates training as only the pose regressor needs to be trained for each new scene. Moreover, the pose regressor's design allows the feature aggregator to be customized for specific scenes, accommodating their unique geometries and objects.

The aggregator incorporates an MLP-Mixer \cite{mlp-mixer} layer for feature mixing, integrating global point relationships, followed by global pooling to generate a single descriptor for the point cloud. This descriptor is then processed by the pose predictor to determine the 6-DoF pose. The regressor is trained using a pose loss \cite{geometricloss}. To ensure robust global descriptors, we implement a projector MLP, whose output is used for metric learning or contrastive loss regularization. This enhancement boosts performance while maintaining rapid training times. To our knowledge, FlashMix is the first to incorporate contrastive regularization in a LiDAR pose regression framework. This innovative approach enables swift adaptation to new environments, making it ideal for real-world scenarios requiring rapid deployment with reduced storage and communication requirements, making it suitable for single and multi-robot localization systems. 

In summary, our contributions are:
\begin{itemize}
    \item A novel map-free localization framework combining a pre-trained point encoder with a scene-specific pose regressor with feature buffer enabled rapid training.
    \item Integration of an MLP-Mixer as a descriptor aggregator, to fuse global point relationships by feature mixing for adapting to scene-specific geometries.
    \item Introduce metric learning and contrastive loss regularization, enhancing global descriptor quality for stable convergence while maintaining fast training times.
    \item Extensive experiments in outdoor and indoor environments, demonstrating rapid training and adaptation with competitive performance compared to existing map-free localization methods.
\end{itemize}

The rest of the paper is organized as follows: Sec. \ref{sec:related-works} covers related work, Sec. \ref{sec:methodology} presents the problem and our framework, Sec. \ref{sec:experiments} shows experimental results, and Sec. \ref{sec:conclusion} concludes the paper.

\section{Related Works}
\label{sec:related-works}
Traditional LiDAR relocalization systems employing maps typically use retrieval and matching-based approaches for pose estimation. These methods process inputs in the form of BEV projections~\cite{bvmatch, scancontext} or raw point clouds~\cite{pointnetvlad,egonn,lcdnet,logg3dnet,salsa} to extract local descriptors. The extraction is achieved either through histogram/statistics-based techniques or learned from data. These frameworks then aggregate these local descriptors to generate a global descriptor, which is used to retrieve nearby point clouds from the map. Subsequently, local descriptors from these retrieved point clouds are matched to estimate pose through 3D registration.
To enhance pose estimation accuracy, multiple candidate point clouds can be retrieved from the map for a given query.  These candidates undergo reranking based on RANSAC registration-derived geometric fitness scores or through spectral methods~\cite{spectralgv}. This multi-step approach yields more robust and precise localization within the mapped environment, albeit at the cost of increased computational complexity and storage requirements.

Map-free localization approaches address these issues by predicting pose directly from the input image or point cloud using regression, avoiding the need for memory-intensive databases and costly registration. Camera pose regression has been extensively researched, with various deep learning methods~\cite{posenet,mapnet,atloc}. Notable works include HSCNet~\cite{hscnet}, which uses regional classification and FiLM conditioning~\cite{film}, and its extension SRCNet~\cite{srcnet} for few-shot learning. ReCoLoc~\cite{recoloc} incorporated region contrastive representation learning, while DSAC*\cite{dsac*} employed scene coordinate regression with differentiable RANSAC\cite{ransac} for end-to-end pose prediction. Recent advancements accelerate training by using uniform~\cite{ace} or guided~\cite{focustune} random shuffling of image patch features from a training buffer to decorrelate gradients for learning scene-specific MLP from scene-agnostic dense feature encoder. However, while these methods perform well on small, camera-captured scenes, they face challenges scaling to large environments.

The pioneering map free localization method, PointLoc \cite{pointloc}, uses a PointNet++ \cite{pointnet++} encoder with self-attention to directly estimate 6-DoF poses from LiDAR frames by minimizng pose loss~\cite{geometricloss}. Building on this, \cite{posepn++} PosePN, PosePN++, PoseSOE, and PoseMinkloc each with different encoders, showing that encoder weights for the same models could be transferred across datasets. STCLoc \cite{stcloc} incorporated spatio-temporal constraints to handle dynamic environments, while NIDALOC \cite{nidaloc} implemented a Hebbian memory module to preserve historical information. Hypliloc \cite{hypliloc} enhanced performance by fusing descriptors from 3D and spherical representations of point clouds using a hyperbolic fusion function. Departing from direct pose prediction, SGLoc \cite{sgloc} adopted scene coordinate regression, predicting 3D scene coordinates for each point cloud using Kabsch \cite{kabsch} algorithm in a RANSAC loop for pose estimation. Recent approaches have explored generative paradigms, with DiffLoc \cite{diffloc} proposing a multi-step inference process using stable diffusion-based denoising for pose prediction. LiSA \cite{lisa} utilizes diffusion-based distillation from a 3D semantic segmentation model to learn a multi-scale feature extractor for scene coordinate regression and subsequent pose estimation. Although while SGLoc, DiffLoc, and LiSA have demonstrated promising results with high localization accuracy, they involve lengthy training times and/or computationally intensive evaluation processes, presenting challenges for rapid deployment.

Our method improves upon existing approaches by using a scene-agnostic point encoder and a scene-specific pose regressor consisting of an MLP-Mixer-based descriptor aggregator and MLP pose predictor. This setup, enhanced with pose loss and metric or contrastive loss-based regularization, directly predicts the pose while incorporating global relationships through Mixer layers for improved aggregation. FlashMix accelerates training by using a training buffer of local point descriptors extracted from scene-agnostic encoder, significantly reducing the computational overhead. This approach allows for rapid adaptation to new environments without compromising accuracy due to the scene-specific aggregator, resulting in a highly competitive performance with significantly reduced training time.

\section{Methodology}
\label{sec:methodology}

\subsection{Problem Statement}
The objective of our map-free LiDAR localization framework is to determine the 6-DoF pose of a LiDAR sensor within a scene from a single LiDAR scan. The LiDAR pose is defined as a rigid transformation that maps coordinates from the LiDAR's local frame to the global scene frame. Our framework, therefore, learns this rigid transformation as a function \(\Phi: \mathbb{R}^{N \times 3} \rightarrow \mathbb{R}^{6}\), which takes a point cloud \(Q \in \mathbb{R}^{N \times 3}\) of \(N\) points as input and outputs its 6-DoF pose in the scene. This pose comprises a 3-dimensional position vector and a 3-dimensional orientation vector, the latter being represented as the logarithm of the unit quaternion.

In line with previous work, we formulate \(\Phi\) as a composite function \(\Phi = g(h(.))\). Here, \(h : \mathbb{R}^{N \times 3} \rightarrow \mathbb{R}^{M \times d}\) is the feature encoder that encodes each point into a feature of dimension \(d\) and downsamples the point cloud from \(N\) to \(M\) points. The function \(g : \mathbb{R}^{M \times d} \rightarrow \mathbb{R}^6\) is the regressor that predicts the 6-DoF pose as described above. In our approach, \(h\) is pre-trained on a large dataset, making it scene-agnostic and not specific to any particular scene. For each new scene, \(h\) is frozen, and only \(g\) is trained.

\subsection{Scene-agnostic Backbone}
\label{sec:encoder}
FlashMix leverages the SALSA \cite{salsa} backbone to encode the input point cloud into a higher-dimensional space, generating robust point descriptors. SALSA's backbone is a SphereFormer \cite{sphereformer} that utilizes a U-Net \cite{unet} backbone with sparse 3D convolutions \cite{sparseconv}, and Spherical Transformer layers at each depth. The Spherical Transformer block combines cubic-window attention with radial-window attention, ensuring attention computation for distant points within the same radial window. SALSA was trained end-to-end on the extensive Mulran \cite{mulran} and Apollo-Southbay \cite{southbay} datasets for LiDAR place recognition. In FlashMix, we use the pre-trained weights of SALSA's backbone and keep it frozen while training the scene-specific pose regressor.

The input LiDAR point cloud is preprocessed by removing the ground plane and voxelized with a voxel size of 0.5m to get \(Q \in \mathbb{R}^{N \times 3}\) which is processed through the backbone to generate descriptors for each point, resulting in an output \(\hat{F} \in \mathbb{R}^{N \times d}\), where \(d\) represents the feature dimension. To manage the variability in the number of points in each point cloud, we apply farthest point sampling (FPS) on \(Q\). This process selects a subset of points and their corresponding descriptors, producing \(F \in \mathbb{R}^{M \times d}\). The FPS algorithm begins by randomly selecting an initial point, then iteratively chooses the point farthest from the already selected points until \(M\) points are chosen. This method ensures better coverage of the entire point cloud compared to random sampling. In our experiments, for outdoor scenes, \(M\) is set to 1024, while for indoor scenes, \(M\) is set to 512.

\subsection{Scene-specific Regressor}
\label{sec:regressor}
A global descriptor is formed by aggregating the point descriptors using the MLP-Mixer~\ref{sec:descriptor-aggregator} architecture. This descriptor is subsequently processed through the Pose Predictor~\ref{sec:pose_predictor} to estimate the LiDAR's 6-DoF pose.

\subsubsection{Descriptor Aggregator}
\label{sec:descriptor-aggregator}
\begin{figure}[b]
    \centering
    \includegraphics[width=\linewidth]{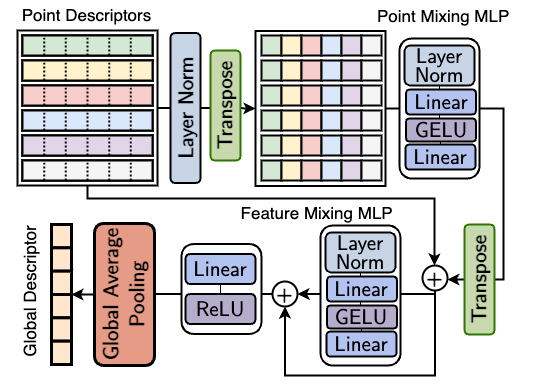}
    \caption{MLP-Mixer Aggregator that fuses local descriptor using point and channel mixing MLPs followed by average pooling.}
    \label{fig:descriptor_aggregator}
\end{figure}

We employ an MLP-Mixer \cite{mlp-mixer} layer to incorporate global relationships within the point descriptors, followed by global average pooling to aggregate a single global descriptor per point cloud (Fig. \ref{fig:descriptor_aggregator}). The MLP-Mixer architecture comprises point-mixing and feature-mixing MLP layers. The point descriptors for a point cloud (\(P_d\) of shape \(M \times d\)) are transposed to shape \(d \times M\) and passed through the point-mixing MLP layer. In this layer, the point descriptors interact with each other, facilitating information sharing and enhancing the global representation. The output is then transposed back to shape \(M \times d\) and processed through the feature-mixing MLP layer. Both the point-mixing and feature-mixing MLPs incorporate layer normalization and two linear layers with GELU \cite{gelu} nonlinearity between them. The output of the feature-mixing MLP is projected to a higher dimension \(l\) using a linear layer with ReLU nonlinearity, followed by global average pooling to obtain an \(l\)-dimensional global descriptor for the point cloud.

\subsubsection{Pose Predictor}
Our training strategy incorporates two key loss components: a pose loss for accurate position and orientation estimation, and a regularization loss to enhance the robustness of global descriptors.
\label{sec:pose_predictor}
The global descriptor \(\in \mathbb{R}^{l}\) is subsequently passed through a series of MLP regressors, each consisting of a linear layer, followed by batch normalization and a ReLU activation. The output is then bifurcated into separate translation and rotation heads, each employing similar MLP structures. The translation head outputs \(t_{pred} \in \mathbb{R}^3\), representing the \(x, y, z\) position of the LiDAR, while the rotation head outputs an orientation \(q_{pred} \in \mathbb{R}^3\), represented as the logarithm of the unit quaternion to avoid singularities.

\subsection{Training Objective}
Our training strategy incorporates two key components: a pose loss for accurate position and orientation estimation, and a metric or contrastive loss-based regularization term to enhance performance by making global descriptors robust.
\subsubsection{Pose Loss}
Following a similar approach as previous work \cite{posepn++, nidaloc}, we implement a pose loss mechanism \cite{geometricloss} that separately applies $L_1$ losses to the predicted position and orientation. The overall pose loss, $\mathcal{L}_{pose}$, is then computed as a weighted sum of these individual losses:
\begin{align}
    \mathcal{L}_{pose} = \|t_{pred} - t_{gt}\|_1 + \alpha \|q_{pred} - q_{gt}\|_1
\end{align}
where \(pred\) denotes the predicted values, \(gt\) denotes the ground truth values, and \(\alpha\) serves as a hyperparameter.
\subsubsection{Contrastive Regularization}
\label{sec:contrastive_reg}
FlashMix incorporates a contrastive regularization component to enhance the robustness and discriminative power of global descriptors. This process begins by projecting global descriptors into a new embedding space using a projector MLP, consisting of two linear layers with a ReLU activation function between them. We apply the Barlow Twins \cite{barlow_twins} contrastive loss on the normalized embeddings. This loss is designed to minimize the distance between embeddings of geometrically close point clouds (positives) while maximizing the distance between embeddings of point clouds that are further apart (negatives). The Barlow Twins contrastive loss, hyperparameter $\mu$ (0.005), is formulated as:
\begin{align}
    \mathcal{L}_{C.L} = \sum_i (1-C_{ii})^2 + \mu \sum_{i}\sum_{j \neq i} C_{ij}^2
\end{align}
where $C$ is the cross-correlation matrix between the descriptors of queries and positives in a batch, and given by
\begin{align}
    C_{i,j} = \frac{\sum_a l^q_{a,i} l^p_{a,j}}{\sqrt{\sum_a (l^q_{a,j})^2} \sqrt{\sum_a (l^p_{a,j})^2}}
    \label{eqn:bt}
\end{align}
where \(l^q_{a,i}\) and \(l^p_{a,i}\) are the values at index $a$ of the projected embeddings of the query (\(l^q_i\)) and its positive counterpart (\(l^p_i\)), respectively. 
This encourages a high correlation between queries and their positives and a low correlation with negatives by minimizing feature redundancy while maximizing invariance to positives, effectively enhancing descriptor discrimination. 

Alternatively, we propose employing triplet margin loss on the projected embeddings as a regularizer, replacing the Barlow Twins contrastive loss. Being typically used in representation learning tasks like place recognition, the triplet margin loss enhances model robustness by ensuring that the Euclidean distance between the embeddings of a query and its positive is smaller than that between the query and a negative. Specifically, for each set of query \(l^q_i\), positive \(l^p_i\), and negative \(l^n_i\), the triplet margin loss is defined as:
\begin{align}
    \mathcal{L}_{M.L.} = \max \left\{ \|l^q_i - l^p_i\|_2^2 - \|l^q_i - l^n_i \|_2^2 + m, 0 \right\}
\end{align}
where the margin \(m\) is set to 0.05. Contrastive loss $\mathcal{L}_{C.L}$ or metric loss $\mathcal{L}_{M.L}$ is composited with the pose loss $\mathcal{L}_{pose}$.

\subsection{FlashMix Training}
The fast training of the pose regressor~\ref{sec:regressor} for each new scene is facilitated by generating a training buffer of point descriptors. This is achieved using the pre-trained backbone, which iterates over the complete dataset of the scene. By employing farthest point sampling, we ensure uniformity in the number of points across all point clouds in the buffer, while also enabling large batch size training on a single GPU. For enhanced training efficiency, we implement mixed precision training and preload the training buffer directly onto the GPU, thereby eliminating communication overhead during data loading. The integration of the MLP-Mixer aggregator and contrastive regularization within FlashMix's training procedure accelerates training times while maintaining high accuracy levels.
\begin{table*}
    \centering
    \begin{tabular}{l? l ?c c c c ?c }
         \Xhline{3\arrayrulewidth}
         \textbf{Methods} & \textbf{Training Time} & \textbf{Full6} & \textbf{Full7} & \textbf{Full8} & \textbf{Full9} & \textbf{Average} \\
         \Xhline{3\arrayrulewidth}
        \textbf{PosePN++} & 590 minutes & 9.59, 1.92 & 10.66, 1.92 & 9.01, 1.51 & 8.44, 1.71 & 9.43, 1.77 \\
        \textbf{NIDALoc} & 1200 minutes & 6.71, 1.33 & 5.45, 1.40 & 6.68, 1.26 & 4.80, 1.18 & 5.91, 1.29 \\
        \textbf{HypLiLoc} & 1020 minutes & 6.00, \textbf{1.31} & 6.88, \textbf{1.09} & 5.82, \textbf{0.97} & 3.45, \textbf{0.84} & 5.54, \textbf{1.05} \\
        \Xhline{3\arrayrulewidth}
        \textbf{FlashMix (ML Reg.)} & \textbf{80 minutes} & 3.15, 2.00 & \textbf{4.07}, 1.88 & \textbf{4.61}, 2.54 & 3.68, 1.79 & 3.88, 2.05 \\
        \textbf{FlashMix (CL Reg.)} & \textbf{80 minutes} & \textbf{3.05}, 1.96 & 4.55, 2.05 & 4.67, 2.05 & \textbf{2.94}, 1.79 & \textbf{3.80}, 1.96 \\
        \Xhline{3\arrayrulewidth}
    \end{tabular}
    \caption{Mean position (m) and orientation errors ($^\circ$) on Oxford-Radar Dataset. Best performance is highlighted in ~\textbf{bold}, lower is better.}
    \label{tab:oxford-results}
\end{table*}

\section{Experiments}
\label{sec:experiments}
\subsection{Dataset and Implementation Details}
We test our framework on three public datasets: Oxford-Radar \cite{oxford}, Mulran DCC \cite{mulran}, and vReLoc \cite{pointloc}. Oxford-Radar and Mulran DCC are large-scale outdoor datasets, while vReLoc is a small indoor dataset. Oxford-Radar dataset captures over 32 traversals in central Oxford, containing sensor data collected over a time span of 1 year and a length span of 1000 km. Moreover, it covers various seasons and weather
conditions. Similar to previous methods~\cite{posepn++, nidaloc, hypliloc}, we trained on the sequences named \textit{2019-01-11-14-02-26, 2019-01-14-12-05-52, 2019-01-14-14-48-55, 2019-01-18-15-20-12} and tested on the sequences named \textit{2019-01-15-13-06-37 (Full6), 2019-01-17-13-26-39 (Full7), 2019-01-17-14-03-00 (Full8), 2019-01-18-14-14-42 (Full9)}. For training our method for the Oxford-Radar dataset, we sample point clouds every 1 meter.
Mulran DCC contains three trajectories of scans collected from an Ouster-64 LiDAR in South Korea. This dataset is more challenging due to multiple trajectory reversals and occlusions. We trained on DCC1 and DCC2 sequences and tested on the DCC3 sequence. vReLoc is an indoor dataset collected inside a room of area 4m x 5m. Several obstacles are laid in the scene. Similar to previous work, we trained on \textit{seq-03, seq-12, seq-15, seq-16} and tested on \textit{seq-05, seq-06, seq-07, seq-14}.

FlashMix was implemented in Pytorch~\cite{pytorch}, and experiments were run on an Nvidia RTX A4000 GPU with an Intel(R) i9 CPU and 128 GB RAM. We used the Adam optimizer with initial learning rates of 0.01 for Oxford-Radar and 0.001 for Mulran DCC and vReLoc datasets. A one-cycle policy with cosine annealing was employed, with final learning rates of \(10^{-6}\) for Oxford-Radar and \(10^{-5}\) for Mulran DCC and vReLoc. Batch sizes were 1024 for Oxford-Radar and Mulran DCC and 1280 for vReLoc.
\subsection{Results}
We compare metric learning (ML Reg.) and contrastive learning (CL Reg.) FlashMix with HypLiLoc~\cite{hypliloc}, NIDALoc~\cite{nidaloc}, and PosePN++~\cite{posepn++} on three datasets described previously. Additional comparisons are in Supp..
\begin{figure}[htbp]
    \centering
    \begin{subfigure}[b]{0.45\textwidth}
        \centering
        \includegraphics[width=\linewidth]{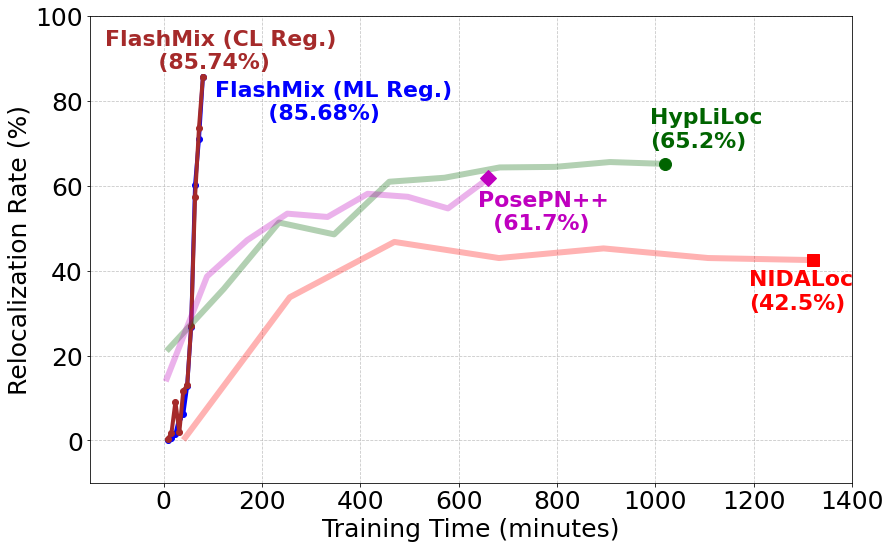}
        \caption{Oxford-Radar}
    \end{subfigure}
    \hfill
    \begin{subfigure}[b]{0.45\textwidth}
        \centering
        \includegraphics[width=\linewidth]{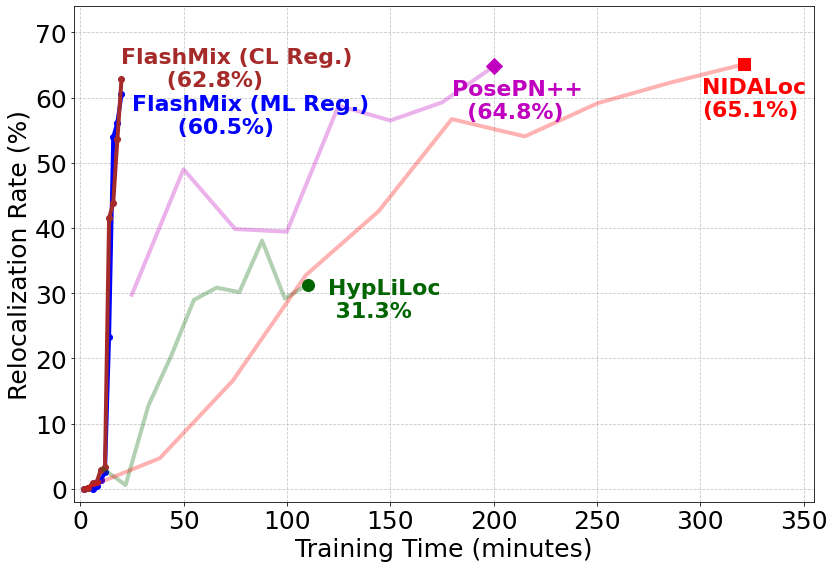}
        \caption{Mulran DCC}
    \end{subfigure}
    \hfill
    \begin{subfigure}[b]{0.45\textwidth}
        \centering
        \includegraphics[width=\linewidth]{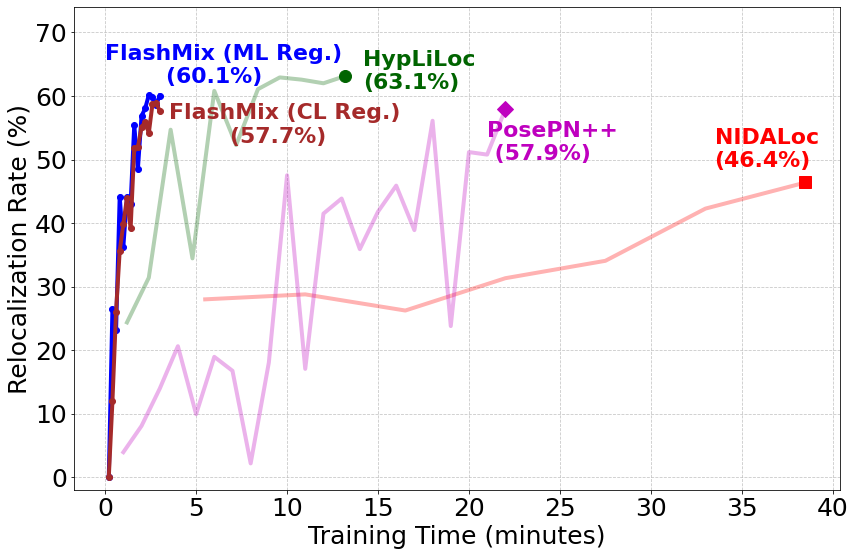}
        \caption{vReLoC}
        \label{fig:res_time_vreloc}
    \end{subfigure}
    \caption{Analysis of relocalization rate as a function of train time.}
    \label{fig:res_time_combined}
\end{figure}

\subsubsection{Training Time vs Relocalization Rate}
Figure ~\ref{fig:res_time_combined} shows relocalization success rates versus training time for three datasets. Relocalization is defined as the percentage of samples within $5$ meters and $5^{\circ}$ degrees for Oxford-Radar and Mulran DCC, and $0.25$ meters and  $5^{\circ}$ for vReLoc. FlashMix achieves competitive results with significantly reduced training times across all datasets. It outperforms HypLiLoc on Oxford-Radar by 20\% with only 1/12th of the training time. On Mulran DCC and vReLoc, FlashMix reaches 62.8\% and 60.1\% relocalization rates (compared to best rates of 65.1\% and 63.1\%), requiring only 20 and 5 minutes of training respectively.
\subsubsection{Translation and Rotation Errors}
Tab. \ref{tab:oxford-results} and \ref{tab:dcc_results} show average translation and rotation errors with training times for Oxford-Radar and Mulran DCC datasets. FlashMix achieves lowest translation errors across all sequences, with FlashMix (CL Reg.) slightly outperforming (ML Reg.). On Oxford-Radar, FlashMix's rotation error (1.96$^\circ$) is higher than HypLiLoc by 0.9$^\circ$ but trains in 80 minutes versus several hours.

Table~\ref{tab:vreloc_results} shows median errors for vReLoc (500 scans/sequence, 5x4m room). FlashMix trains in 5 minutes with low errors, being worse compared to HypLiLoc by 0.04m and 0.9° which requires double the training time.

Mulran DCC's multiple trajectory retraversals and occlusions pose challenges for relocalization due to the point cloud being projected as range images have uninformative pixels manifesting as occlusions. Thus, HypLiLoc struggles with high translation errors whereas FlashMix avoids overfitting, achieving lowest translation error of 5.82 meters.
\begin{table}
\centering
    \begin{tabular}{l ?c| c }
        \Xhline{3\arrayrulewidth}
         \textbf{Methods} & \textbf{Training Time} & \textbf{DCC3} \\
        \Xhline{3\arrayrulewidth}
        \textbf{PosePN++} & 200 mins.& 6.64, 3.43 \\
        \textbf{NIDALoc} & 321 mins.& 5.87, 3.39 \\
        \textbf{HypLiLoc} & 110 mins.& 10.86, \textbf{2.88} \\
        \Xhline{3\arrayrulewidth}
        \textbf{FlashMix (ML Reg.)}& \textbf{20 mins.}& 6.07, 4.17 \\
        \textbf{FlashMix (CL Reg.)}& \textbf{20 mins.}& \textbf{5.82}, 3.96 \\
        \Xhline{3\arrayrulewidth}
    \end{tabular}
    \caption{Mean position (m) and orientation errors ($^\circ$) on DCC.}
    \label{tab:dcc_results}
\end{table}
\begin{table*}
\centering
    \begin{tabular}{ l ? c ? c c c c ?c }
       \Xhline{3\arrayrulewidth}
         \textbf{Methods}& \textbf{Training Time} & \textbf{Seq-05} & \textbf{Seq-06} & \textbf{Seq-07} & \textbf{Seq-14} & \textbf{Average} \\
        \Xhline{3\arrayrulewidth}
        \textbf{PosePN} & 40 minutes & 0.12, 4.38 & 0.09, 3.16 & 0.17, 3.94 & \textbf{0.08}, 3.27 & 0.12, 3.69 \\
        \textbf{PosePN++} & 22 minutes & 0.15, 3.12 & 0.10, 3.31 & 0.15, 2.92 & 0.10, 2.80 & 0.13, 3.04 \\
        \textbf{NIDALoc} & 38 minutes & 0.18, 3.63 & 0.15, 4.09 & 0.21, 3.24 & 0.17, 3.98 & 0.18, 3.74 \\
        \textbf{HypLiLoc} & 13 minutes & \textbf{0.09, 2.52} & \textbf{0.08, 2.58} & \textbf{0.13, 2.55} & 0.09, \textbf{2.34} & \textbf{0.10, 2.50} \\
        \Xhline{3\arrayrulewidth}
        \textbf{FlashMix (ML Reg.)} & \textbf{5 minutes} & 0.14, 3.03 & 0.12, 3.58 & 0.18, 3.70 & 0.11, 3.11 & 0.14, 3.34 \\
        \textbf{FlashMix (CL Reg.)} & \textbf{5 minutes} & 0.16, 3.14 & 0.12, 3.30 & 0.18, 3.92 & 0.11, 3.32 & 0.14, 3.42 \\
        \Xhline{3\arrayrulewidth}
    \end{tabular}
    \caption{Median translation and rotation errors (m/$^\circ$) on the vReLoc dataset.}
    \label{tab:vreloc_results}
\end{table*}

\subsection{Qualitative Comparison}
\begin{figure*}[htbp]
    \centering
    \begin{subfigure}[b]{\textwidth}
        \centering
        \includegraphics[width=0.93\linewidth]{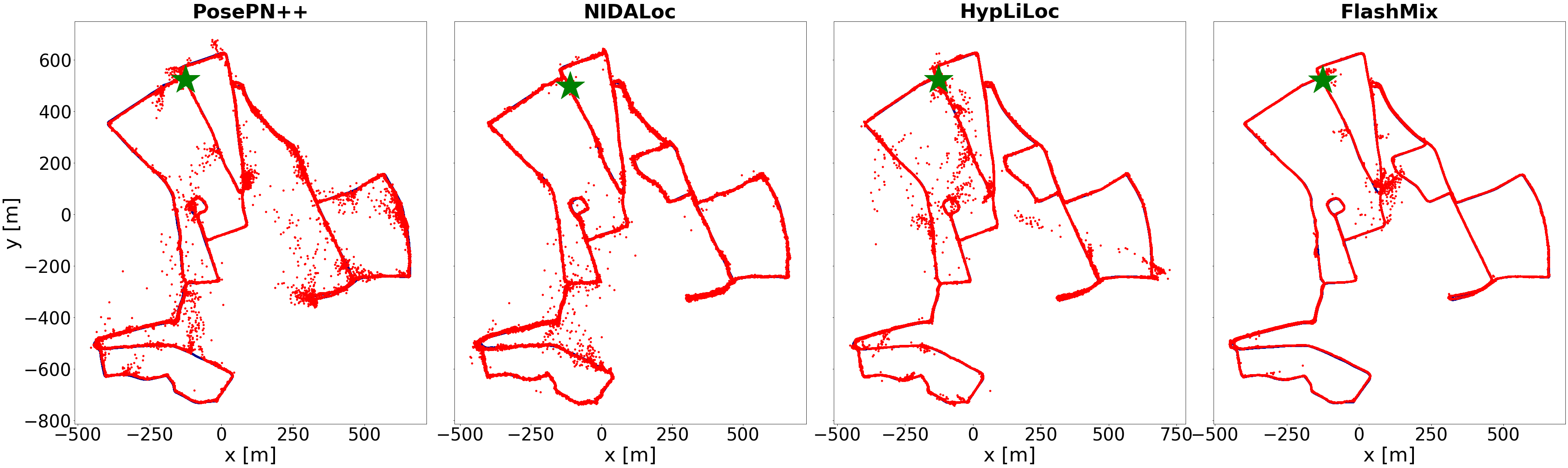}
        \caption{Oxford-Radar Full8}
    \end{subfigure}
    \vspace{0.1cm}
    
    \begin{subfigure}[b]{\textwidth}
        \centering
        \includegraphics[width=0.93\linewidth]{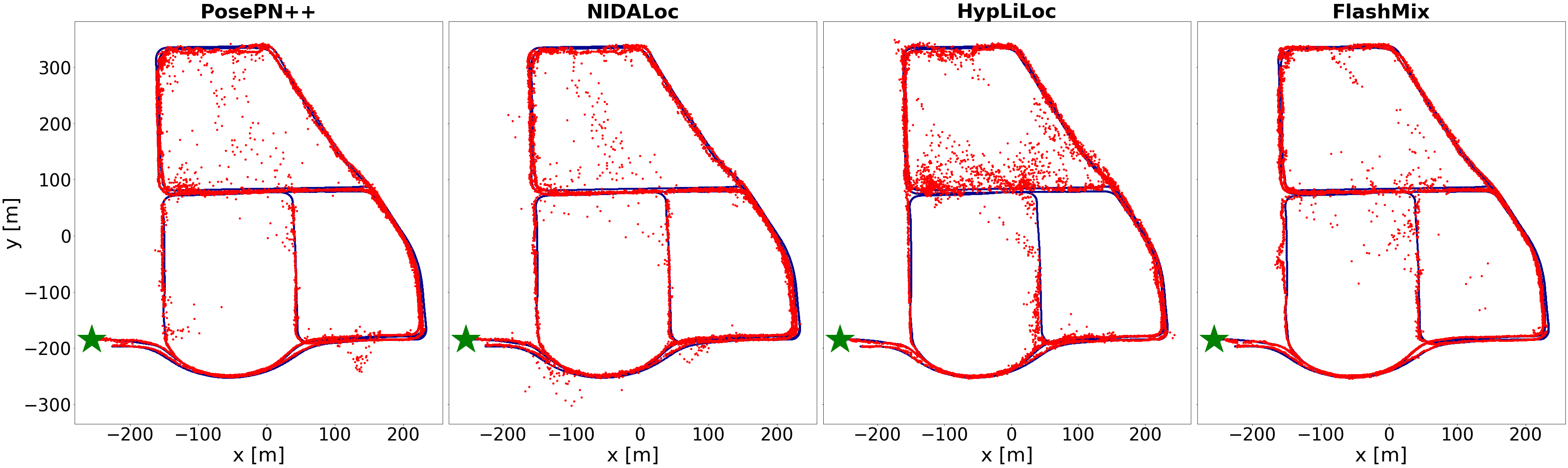}
        \caption{Mulran DCC DCC3}
    \end{subfigure}
    \vspace{0.1cm}
    
    \begin{subfigure}[b]{\textwidth}
        \centering
        \includegraphics[width=0.93\linewidth]{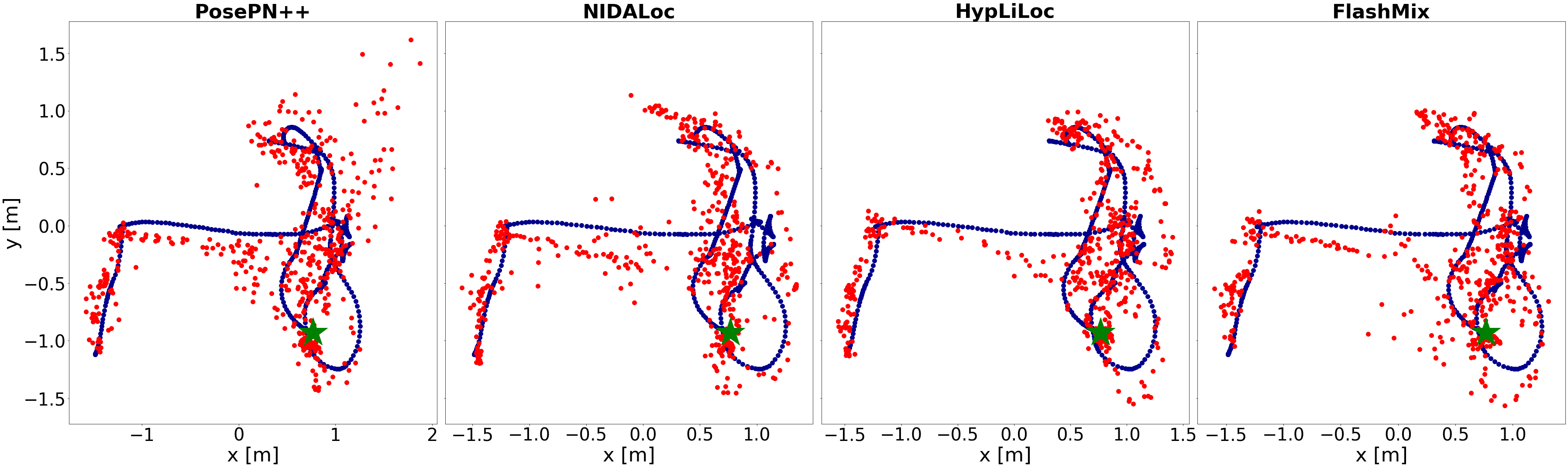}
        \caption{vReLoC seq-05}
        \label{fig:scatter_vreloc}
    \end{subfigure}
    \caption{Visualization of different methods on test trajectories from Oxford-Radar, DCC, and vReLoC dataset. Trajectory visualization: The ground truth and estimated positions are shown in dark blue and red dots, respectively. The star shows the starting position.}
    \label{fig:qualitative-plots}
\end{figure*}

The predicted 2D positions by each method (red), along with the actual values (dark blue), are illustrated in Figure~\ref{fig:qualitative-plots}. We include plots for the Full8 trajectory of the Oxford-Radar dataset, the DCC3 trajectory of the Mulran DCC dataset, and the seq-06 trajectory for the vReLoc dataset. Our method consistently demonstrates high accuracy in prediction, with the predicted positions closely aligning with the ground truth. In contrast, other methods exhibit a greater dispersion of values, particularly in the outdoor datasets.

\subsection{Ablation Studies}
\noindent \textbf{Contrastive Regulariazation}:
We conduct ablation studies with different contrastive losses as training regularizers. Namely, we compare the contrastive losses of SigLIP \cite{siglip}, NTXent \cite{ntxent}, and Barlow Twins \cite{barlow_twins}. We also compare these with the metric learning Triplet Loss because of its use in common LiDAR place recognition frameworks. For the ablation studies, we use the Oxford-Radar dataset and present the relocalization rates in each of its sequences: Full6 (F6), Full7 (F7), Full8 (F8), and Full9 (F9).  Further, the performance of the method without using any contrastive or metric loss regularization is shown for reference. As observed in Tab.~\ref{tab:loss_ablation}, not using any regularization loss resulted in the poorest performance. While NTXent offers higher performance, its quadratic scaling with batch size creates efficiency challenges. To balance performance and efficiency, we use Triplet loss and Barlow Twins loss.
\begin{table}
\centering
    \setlength{\tabcolsep}{3pt}
    \begin{tabular}{ l ? c  |c  |c  |c ? c }
        \Xhline{3\arrayrulewidth}
         & \textbf{F6} & \textbf{F7} & \textbf{F8} & \textbf{F9} & \textbf{Avg.} \\
        \Xhline{3\arrayrulewidth}
        \textbf{No Reg. Loss} & 88.92 & 78.15 & 76.32 & 89.72 & 82.77 \\
        \textbf{SigLIP} & 88.14 & 81.01 & 79.43 & 90.87 & 84.48 \\
        \textbf{NTXent} & 88.63 & \textbf{82.29} & \textbf{81.58} & \textbf{92.56} & \textbf{85.92} \\
        \textbf{Triplet} & \textbf{91.95} & 82.13 & 78.80 & 91.80 & 85.69 \\
        \textbf{Barlow Twins} & 91.82 & 81.56 & 80.42 & 90.92 & 85.74 \\
        \Xhline{3\arrayrulewidth}
    \end{tabular}
\caption{Ablation Study: Impact of Contrastive and Metric Loss regularization.}
\label{tab:loss_ablation}
\end{table}

\noindent \textbf{Descriptor Aggregator}:
In this ablation study, we explored the impact of different aggregator blocks on FlashMix's performance. Namely, we experimented with Mixer+Global Average Pooling (GAP), MLP+GAP, Multi-headed Attention+GAP, and Mixer+SALAD~\cite{salad}. The performance of these aggregators, when combined with Barlow Twins regularization and Triplet Regularization, is detailed in Tab. \ref{tab:agg_barlow} and \ref{tab:agg_triplet}, respectively. With Triplet Regularization, Mixer+GAP achieved the best results while also requiring less training time. Under Barlow Twins regularization, although Mixer+SALAD performed the best, it required approximately 135 minutes of training, compared to the Mixer+GAP, which achieved slightly lower results but needed only 80 minutes. Consequently, we adopted the Mixer+GAP architecture for all our experiments.

\begin{table}
    \centering
    \setlength{\tabcolsep}{1.5pt}
    \begin{tabular}{ l ? l ? c | c | c | c ? c }
        \Xhline{3\arrayrulewidth}
        \textbf{Aggregator} & \textbf{Time} & \textbf{F6} & \textbf{F7} & \textbf{F8} & \textbf{F9} & \textbf{Avg.} \\
        \Xhline{3\arrayrulewidth}
        \textbf{MLP + GAP} & \textbf{70}  & 89.26 & 80.52 & 78.68 & 90.50 & 84.31 \\
        \textbf{MHA + GAP} & 225 & 78.37 & 65.12 & 62.79 & 86.93 & 72.55 \\
        \textbf{Mixer+SALAD} & 135 & 90.20 & 79.00 & 78.81 & 91.01 & 84.27 \\
        \textbf{Mixer+GAP} & 80  & \textbf{91.95} & \textbf{82.13} & \textbf{78.80} & \textbf{91.80} & \textbf{85.69} \\
        \Xhline{3\arrayrulewidth}
    \end{tabular}
    \caption{Ablation Study: Descriptor Aggregators with Triplet loss regularization. Here, time refers to training time in minutes.}
    \label{tab:agg_triplet}
\end{table}

\begin{table}
    \centering
    \setlength{\tabcolsep}{1.5pt}
    \begin{tabular}{ l ? l | c | c | c | c ? c }
    \Xhline{3\arrayrulewidth}
    \textbf{Aggregator} & \textbf{Time} & \textbf{F6} & \textbf{F7} & \textbf{F8} & \textbf{F9} & \textbf{Avg.} \\
    \Xhline{3\arrayrulewidth}
    \textbf{MLP + GAP} & \textbf{70} & 88.06 & 80.64 & 81.37 & 89.96 & 84.68 \\
    \textbf{MHA + GAP} & 225 & 65.17 & 62.23 & 58.98 & 84.34 & 67.14 \\
    \textbf{Mixer+SALAD} & 135  & \textbf{93.67} & \textbf{82.50} & \textbf{83.47} & \textbf{93.56} & \textbf{87.86} \\
    \textbf{Mixer+GAP} & 80  & 91.82 & 81.56 & 80.42 & 90.92 & 85.74 \\
    \Xhline{3\arrayrulewidth}
    
    \end{tabular}
    \caption{Ablation Study: Descriptor Aggregators with Barlow Twins regularization. Time refers to training time in minutes.}
    \label{tab:agg_barlow}

\end{table}

\noindent \textbf{Descriptor Dimension}: Our findings indicated that increasing the descriptor dimension consistently enhanced performance but also lengthened the training time, as shown in Table~\ref{tab:ablation_desc}. To balance performance gains with training efficiency, we settled on a descriptor dimension of 1024.

\begin{table}
    \centering
    \setlength{\tabcolsep}{1.5pt}
    \begin{tabular}{ l ? l ? c | c | c | c ? c }
        \Xhline{3\arrayrulewidth}
        \textbf{No. of Layers} & \textbf{Time} & \textbf{F6} & \textbf{F7} & \textbf{F8} & \textbf{F9} & \textbf{Avg.} \\
        \Xhline{3\arrayrulewidth}
        \textbf{1} & \textbf{80}  & \textbf{91.95} & 82.13 & 78.80 & \textbf{91.80} & 85.69 \\
        \textbf{2} & 90  & 91.60 & 82.28 & 80.11 & 90.49 & 85.70 \\
        \textbf{3} & 100  & 91.83 & 82.07 & \textbf{81.28} & 90.90 & 86.11 \\
        \textbf{4} & 115  & 91.60 & \textbf{83.09} & 80.6 & 91.21 & \textbf{86.22} \\
        \Xhline{3\arrayrulewidth}
    
    \end{tabular}
    \caption{Ablation Study: Number of Mixer layers. Here, time refers to training time in minutes.}
    \label{tab:ablation_mixer}

\end{table}

\noindent\textbf{Aggregator and Pose Predictor layer depths} While adding more mixer layers generally improved performance, it also resulted in longer training times, as detailed in Table~\ref{tab:ablation_mixer}. Consequently, we opted for a single mixer layer in FlashMix to optimize efficiency. Additionally, we chose six layers for the pose predictor as the performance saturates after about 6 layers (Table~\ref{tab:predictor_layers}).

\begin{table}
    \centering
    \setlength{\tabcolsep}{1.5pt}
    \begin{tabular}{ l ? l ? c | c | c | c ? c }
        \Xhline{3\arrayrulewidth}
        \textbf{Desc. Dim.} & \textbf{Time} & \textbf{F6} & \textbf{F7} & \textbf{F8} & \textbf{F9} & \textbf{Avg.} \\
        \Xhline{3\arrayrulewidth}
        \textbf{256} & \textbf{75}  & 88.13 & 75.62 & 71.2 & 87.64 & 80.03 \\
        \textbf{512} & 80  & 88.57 & 77.85 & 76.02 & 89.45 & 82.47 \\
        \textbf{1024} & 80  & 91.95 & 82.13 & 78.80 & 91.80 & 85.69 \\
        \textbf{2048} & 110  & \textbf{92.46} & \textbf{83.55} & \textbf{82.00} & \textbf{92.18} & \textbf{87.146} \\
        \Xhline{3\arrayrulewidth}
    
    \end{tabular}
    \caption{Ablation Study: Global Descriptor Dimension. Here, time refers to training time in minutes.}
    \label{tab:ablation_desc}

\end{table}

\begin{table}
    \centering
    \setlength{\tabcolsep}{1.5pt}
    \begin{tabular}{ l ? l ? c | c | c | c ? c }
        \Xhline{3\arrayrulewidth}
        \textbf{No. of Layers} & \textbf{Time} & \textbf{F6} & \textbf{F7} & \textbf{F8} & \textbf{F9} & \textbf{Avg.} \\
        \Xhline{3\arrayrulewidth}
        \textbf{4} & 80  & 84.67 & 76.1 & 74.25 & 87.48 & 80.17 \\
        \textbf{6} & 80  & \textbf{91.95} & 82.13 & 78.8 & 91.80 & 85.69 \\
        \textbf{8} & 80  & 91.76 & \textbf{82.71} & \textbf{80.96} & \textbf{91.81} & \textbf{86.39} \\
        \Xhline{3\arrayrulewidth}
    
    \end{tabular}
    \caption{Ablation Study: Number of Pose predictor layers. Here, time refers to training time in minutes.}
    \label{tab:predictor_layers}

\end{table}

\section{Conclusion}
\label{sec:conclusion}
FlashMix addresses the challenge of long training times in map-free LiDAR localization systems while maintaining accuracy and studies the effect of contrastive/metric regularization on pose estimation performance. It utilizes a frozen, scene-agnostic backbone, a descriptor buffer, and an MLP mixer with contrastive or metric loss regularization to rapidly adapt to new scenes. Our extensive evaluations across various LiDAR localization benchmarks demonstrate FlashMix's effectiveness in delivering fast and accurate localization in real-world scenarios.

\clearpage

{\small
\bibliographystyle{ieee_fullname}
\bibliography{egbib}
}

\clearpage
\renewcommand\thesection{\Alph{section}}
\setcounter{section}{0}  
\section*{Supplementary Material}
\section{Translation and Rotation Errors}
In Section 4.2 of the manuscript, we evaluated FlashMix against the leading LiDAR pose regression methods of HypLiLoc~\cite{hypliloc}, NIDALoc~\cite{nidaloc}, and PosePN++~\cite{posepn++}. Now we show results from additional methods like PosePN, PoseSOE, PoseMinkLoc~\cite{posepn++}, and PointLoc~\cite{pointloc} for the Oxford-Radar and vReLoc datasets in Tables~\ref{tab:supp-oxford-results} and \ref{tab:supp-vreloc-results}, respectively. Results from retrieval-based methods such as PointNetVLAD~\cite{pointnetvlad} and DCP~\cite{dcp} are also presented for the Oxford-Radar dataset to provide a broader performance context. FlashMix demonstrates the lowest translation errors on the Oxford-Radar dataset and exhibits competitive performance on the vReLoc dataset, all while requiring significantly less training time.

\begin{table*}
\centering

    \begin{tabular}{l? c ?c c c c ?c }
        \Xhline{3\arrayrulewidth}
        \textbf{Method} & \textbf{Training Time} & \textbf{Full6} & \textbf{Full7} & \textbf{Full8} & \textbf{Full9} & \textbf{Average} \\
        \Xhline{3\arrayrulewidth}
        \textbf{PNVLAD} & - & 18.14, 3.28 & 24.57, 3.08 & 19.93, 3.13 & 15.59, 2.63 & 19.56, 3.03 \\
        \textbf{DCP} & - & 16.04, 4.54 & 16.22, 3.56 & 14.87, 3.45 & 12.97, 3.99 & 15.03, 3.89 \\
        \textbf{PosePN} & - & 14.32, 3.06 & 16.97, 2.49 & 13.48, 2.60 & 9.14, 1.78 & 13.48, 2.48 \\
        \textbf{PoseSOE} & - & 7.59, 1.94 & 10.39, 2.08 & 9.21, 2.12 & 7.27, 1.87 & 8.62, 2.00 \\
        \textbf{PoseMinkLoc} & - & 11.20, 2.62 & 14.24, 2.42 & 12.35, 2.46 & 10.06, 2.15 & 11.96, 2.41 \\
        \textbf{PointLoc} & - & 12.42, 2.26 & 13.14, 2.50 & 12.91, 1.92 & 11.31, 1.98 & 12.45, 2.17 \\
        \textbf{PosePN++} & 590 minutes & 9.59, 1.92 & 10.66, 1.92 & 9.01, 1.51 & 8.44, 1.71 & 9.43, 1.77 \\
        \textbf{NIDALoc} & 1200 minutes & 6.71, 1.33 & 5.45, 1.40 & 6.68, 1.26 & 4.80, 1.18 & 5.91, 1.29 \\
        \textbf{HypLiLoc} & 1020 minutes & 6.00, \textbf{1.31} & 6.88, \textbf{1.09} & 5.82, \textbf{0.97} & 3.45, \textbf{0.84} & 5.54, \textbf{1.05} \\
        \Xhline{3\arrayrulewidth}
        \textbf{Flash-Mix (M.L. Reg.)} & \textbf{80 minutes} & 3.153, 2.002 & \textbf{4.066}, 1.882 & \textbf{4.611}, 2.536 & 3.68, 1.791 & 3.878, 2.053 \\
        \textbf{Flash-Mix (C.L. Reg.)} & \textbf{80 minutes} & \textbf{3.048}, 1.959 & 4.551, 2.049 & 4.674, 2.052 & \textbf{2.943}, 1.791 & \textbf{3.804}, 1.963 \\
        \Xhline{3\arrayrulewidth}
    
    \end{tabular}
    \caption{Mean position (m) and orientation errors ($^\circ$) on Oxford-Radar Dataset. Best performance is highlighted in ~\textbf{bold}, lower is better.}
    \label{tab:supp-oxford-results}

\end{table*}

\begin{table}
\centering

    \begin{tabular}{l? c ?c }
        \Xhline{3\arrayrulewidth}
        \textbf{Methods} & \textbf{Training Time} & \textbf{Average} \\
        \Xhline{3\arrayrulewidth}
        \textbf{PosePN} & 40 minutes & 0.12, 3.69 \\
        \textbf{PoseSOE} & - & 0.13, 3,08 \\
        \textbf{PoseMinkLoc} & - & 0.15, 4.57 \\
        \textbf{PointLoc} & - & 0.12, 3.07 \\
        \textbf{PosePN++} & 22 minutes & 0.13, 3.04 \\
        \textbf{NIDALoc} & 38 minutes & 0.18, 3.74 \\
        \textbf{HypLiLoc} & 13 minutes & \textbf{0.10, 2.50} \\
        \Xhline{3\arrayrulewidth}
        \textbf{Flash-Mix (ML Reg.)} & \textbf{5 minutes} & 0.14, 3.34 \\
        \textbf{Flash-Mix (CL Reg.)} & \textbf{5 minutes} & 0.14, 3.42 \\
        \Xhline{3\arrayrulewidth}
    
    \end{tabular}
    \caption{Average of the Median position (m) and orientation errors ($^\circ$) on vReLoc sequences. Best performance is highlighted in ~\textbf{bold}, lower is better.}
    \label{tab:supp-vreloc-results}

\end{table}

\section{Contrastive Regularization}
In the manuscript, we demonstrated how integrating contrastive regularization enhances FlashMix's efficacy. Specifically, we assessed FlashMix's performance with the inclusion of the contrastive regularization losses of SigLIP~\cite{siglip}, NTXent~\cite{ntxent}, and Barlow Twins~\cite{barlow_twins}, alongside the metric-based Triplet Loss. Here, we provide more details into these losses.

SigLIP is a contrastive loss defined as:
\begin{align}
    \mathcal{L}_{SigLIP} = -\frac{1}{|B|}\sum_{i=1}^{|B|}\sum_{j=1}^{|B|} \log{\frac{1}{1+e^{z_{ij}(-tl_i^q.l_j^p + b)}}}
\end{align}
where $l_i^q$ is the query instance at index $i$ in the batch, $l_j^p$ is the positive to the query instance at index $j$, respectively, $|B|$ represents the batch size, $z_{ij}=1$ when $i=j$ and $z_{ij}=-1$ when $i\neq j$. The parameters $t$ (temperature) and $b$ (bias) govern the loss scaling and offset, respectively. Following common practice~\cite{siglip}, the temperature $t$ is parameterized as $\exp{(\bar{t})}$, with $\bar{t}$ being a trainable parameter initially set to $\log{\frac{1}{0.07}}$, and the trainable bias $b$ starting at 0.

For query $l_i^q$ and its positive $l_i^p$, the NTXent (Normalized Temperature-Scaled Cross-Entropy) Loss~\cite{ntxent} is defined as 
\begin{align}
     &\mathcal{L}_{i,j} = -\log\frac{\exp\left(\text{sim}\left(l_i^q, l_j^p\right)/\tau\right)}{\sum_{k}1_{[k\neq{i}]}\exp\left(\text{sim}\left(l_i^q, l_k^p\right)/\tau\right)} \\
     &\mathcal{L}_{NTXent} = \sum_{i,j} \mathcal{L}_{i,j}
\end{align}
where $1_{[k\neq{i}]}$ is the indicator function, which is 1 if $k\neq i$, and 0 otherwise.
The function $\text{sim}(l_i^q, l_i^p)$ calculates the cosine similarity between vectors $l_i^q$ and $l_i^p$, and $\tau$ is a temperature parameter set to 0.07.

The Barlow Twins contrastive loss, with hyperparameter $\mu$ (0.005), is formulated as:
\begin{align}
    \mathcal{L}_{BarlowTwins} = \sum_i (1-C_{ii})^2 + \mu \sum_{i}\sum_{j \neq i} C_{ij}^2
\end{align}
where $C$ is the cross-correlation matrix between the descriptors of queries and positives in a batch, and given by
\begin{align}
    C_{i,j} = \frac{\sum_a l^q_{a,i} l^p_{a,j}}{\sqrt{\sum_a (l^q_{a,j})^2} \sqrt{\sum_a (l^p_{a,j})^2}}
\end{align}
where \(l^q_{a,i}\) and \(l^p_{a,i}\) are the values at index $a$ of the projected embeddings of the query (\(l^q_i\)) and its positive counterpart (\(l^p_i\)), respectively. 

For each set of query \(l^q_i\), positive \(l^p_i\), and negative \(l^n_i\), the triplet margin loss is defined as:
\begin{align}
    \mathcal{L}_{TripletLoss} = \max \left\{ \|l^q_i - l^p_i\|_2^2 - \|l^q_i - l^n_i \|_2^2 + m, 0 \right\}
\end{align}
where the margin \(m\) is set to 0.05.

The relocalization success rate comparison while using contrastive and metric loss regularization is shown in Table~\ref{tab:supp_loss_ablation} (Table 4 of the manuscript). Not using any regularization loss resulted in the poorest performance. 
Among the contrastive loss methods, NTXent achieved the highest average relocalization rate at 85.92\%, closely followed by Barlow Twins with a rate of 85.74\%. Meanwhile, the metric-learning-based Triplet Loss posted a rate of 85.69\%.

While NTXent demonstrates higher performance, its computational cost scales quadratically with the batch size, posing significant efficiency challenges. In contrast, the computational cost for Barlow Twins scales linearly, which substantially reduces training times. Consequently, to optimize the balance between performance and computational efficiency, we integrated Barlow Twins Contrastive regularization into FlashMix. Additionally, we developed a variant of FlashMix utilizing Triplet Loss regularization, thereby offering two distinct configurations tailored to different operational needs. 

\begin{table}
\centering
    \setlength{\tabcolsep}{3pt}
    \begin{tabular}{ l ? c  |c  |c  |c ? c }
        \Xhline{3\arrayrulewidth}
         & \textbf{F6} & \textbf{F7} & \textbf{F8} & \textbf{F9} & \textbf{Avg.} \\
        \Xhline{3\arrayrulewidth}
        \textbf{No Reg. Loss} & 88.92 & 78.15 & 76.32 & 89.72 & 82.77 \\
        \textbf{SigLIP} & 88.14 & 81.01 & 79.43 & 90.87 & 84.48 \\
        \textbf{NTXent} & 88.63 & \textbf{82.29} & \textbf{81.58} & \textbf{92.56} & \textbf{85.92} \\
        \textbf{Triplet} & \textbf{91.95} & \uline{82.13} & 78.80 & \uline{91.80} & 85.69 \\
        \textbf{Barlow Twins} & \uline{91.82} & 81.56 & \uline{80.42} & 90.92 & \uline{85.74} \\
        \Xhline{3\arrayrulewidth}
    \end{tabular}
\caption{Ablation Study: Impact of Contrastive and Metric Loss regularization. The best and second best performances are highlighted in \textbf{bold} and \uline{underline}, respectively.}
\label{tab:supp_loss_ablation}
\end{table}


\section{Descriptor Aggregator}
Section 4.4 of the manuscript explores various descriptor aggregation techniques, including MLP+Global Average Pooling (GAP), Multi-headed Attention (MHA)+GAP, Mixer+SALAD, and Mixer+GAP. Below, we detail each method used in our ablation studies:

\noindent\textbf{MLP+GAP}: This approach utilizes a Multilayer Perceptron (MLP) that features a linear layer followed by a ReLU nonlinearity. The point descriptors are projected to the global descriptor dimension and subsequently processed via Global Average Pooling to yield a singular global descriptor for each point cloud.

\noindent\textbf{MHA+GAP}: This method employs a transformer architecture with multi-headed attention, followed by GAP, for descriptor aggregation. The transformer configuration includes four attention heads, facilitating intricate interactions among point descriptors within each point cloud.

\noindent\textbf{Mixer+SALAD}: The Sinkhorn Algorithm for Locally Aggregated Descriptors (SALAD) technique refines the NetVLAD framework for feature-to-cluster assignment using an optimal transport mechanism. SALAD processes point features through the optimal transport block and integrates the output with a global token to construct robust global descriptors. Although this configuration demonstrated higher performance with Barlow Twins loss in Table 6 of our manuscript, its computational intensity restricted batch sizes to smaller numbers, consequently extending training times.

\noindent\textbf{Mixer+GAP}: This setup, which is the standard across all our experiments as discussed in Section 3.3.1, combines a Mixer with GAP to form the descriptor aggregator.


\end{document}